\documentclass[smallabstract,smallcaptions]{dccpaper}

\usepackage{epsfig}
\usepackage{amsmath,amssymb,amsfonts,amsthm,mathtools}
\usepackage{bm}
\usepackage{color}
\usepackage{url}

\usepackage{times}
\usepackage{graphicx}
\usepackage{wrapfig}
\usepackage{booktabs}
\usepackage{multirow}
\usepackage{tabularx}
\usepackage{array}
\usepackage{float}
\usepackage{xcolor}
\usepackage{soul}
\sethlcolor{yellow}

\usepackage{enumitem}

\usepackage{algorithm}
\usepackage{algpseudocode}

\usepackage{textcomp}
\usepackage{microtype}
\usepackage{nicefrac}

\usepackage[pagebackref,breaklinks,colorlinks]{hyperref}

\newcommand{\fallnest}{\textsc{FallNest}}
\newcommand{\T}{\mathcal{T}}
\newcommand{\Sset}{\mathcal{S}}
\newcommand{\V}{\mathcal{V}}
\newcommand{\Ttest}{\mathcal{T}_{\mathrm{test}}}
\newtheorem{corollary}{Corollary}

\newlength{\figurewidth}
\newlength{\smallfigurewidth}
\begin{document}

\title
{\large
\textbf{\fallnest{}: Dual-Domain Dataset Condensation of Imbalanced IMU Data for Wearable Fall Detection}
}

\author{%
Rashi Niyas P$^{\ast}$, Dr. Ritika Jain$^{\dag}$, and Dr. Hitika Tiwari$^{\ast}$\\[0.5em]
{\small\begin{minipage}{\linewidth}\begin{center}
\begin{tabular}{ccc}
$^{\ast}$Indian Institute of Technology Madras & \hspace*{0.5in} & $^{\dag}$Indian Institute of Technology Guwahati \\
Zanzibar && Assam, India \\
zda24m005@iitmz.ac.in, hitika@iitmz.ac.in && rjain1992@iitg.ac.in
\end{tabular}
\end{center}\end{minipage}}
}

\maketitle
\thispagestyle{empty}

\begin{abstract}
Wearable fall detection is hampered by rare fall events, scarce labeled IMU data, and the cost of training deep models on large, imbalanced corpora. Existing methods either rely on hand-crafted features with shallow classifiers, which limits learning from raw signals, or train deep networks on full datasets that require re-sampling to counter class skew and demand substantial data and compute. To address this, we propose a novel \textbf{Fall} detection framework via \textbf{N}ormalized-gradient and \textbf{E}mbedding-matched \textbf{S}ynthetic \textbf{T}ime--frequency condensation, dubbed \textbf{\fallnest{}}, a data-centric approach that distills large, imbalanced IMU datasets into tiny, class-balanced synthetic training sets. Building on dual-domain time-series condensation (CondTSC), \fallnest{} introduces a fall-specific 7-channel input with accelerometer vector magnitude, class-balanced K-Means initialization, and normalized gradient matching with gradient clipping, combined with MMD-based embedding alignment and multi-view Fourier-domain augmentation, so that class balance arises from the condensed set itself rather than from re-sampling. We evaluate on SisFall, KFall, and MobiFall using strict subject-wise splits and naturally imbalanced test sets, reporting accuracy, Macro-F1, and PR-AUC over five seeds. A 102K-parameter TinyCNN trained on only 32 to 146 synthetic windows (about only 1\% of the training windows)  reaches 93.5-95.8\% mean accuracy and 0.89-0.98 PR-AUC, with standard deviations of 1.0-1.7\% in accuracy, demonstrating the efficacy of our approach.
\end{abstract}

\vspace{1em}
\noindent\textbf{Keywords:} Fall detection, wearable sensors, dataset condensation, class imbalance, dual-domain matching.

\Section{Introduction}
\label{sec:intro}

Falls are a leading cause of injury among older adults and a recurring hazard in workplaces~\cite{ref:who_falls}, and wearable inertial measurement units (IMUs) enable continuous, unobtrusive monitoring~\cite{ref:wearable_review}. Learning a reliable fall detector from IMU data is hard for two reasons. Falls are rare compared with activities of daily living (ADLs), so labeled data are scarce and skewed~\cite{ref:imbalance_challenge}, and wearables operate under tight power, memory, and processing budgets~\cite{ref:embedded_ml_constraints}. The usual remedies for skew, oversampling and augmentation, enlarge the training set and its compute cost. Lightweight pipelines instead rely on hand-crafted features, for example spectral features from 25\,Hz IMU data fed to a feed-forward network~\cite{ref:campanella2024}, which limits learning from raw signals. Deep networks learn features automatically and reach near-perfect accuracy on standard benchmarks~\cite{zhang2024effective,ahirwar2025optimized,owusu2025litefallnet}, but need large imbalanced training sets, substantial compute, and often re-sampling. Dataset condensation distills a large dataset into a small synthetic set that retains its training value~\cite{ref:dc_intro}. It has mainly been studied on images, and its time-series extension~\cite{ref:condtsc_paper} was not evaluated on wearable fall detection under subject-wise splits with naturally imbalanced test sets.
\\ 
We propose \textbf{Fall} detection framework via \textbf{N}ormalized-gradient and \textbf{E}mbedding-matched \textbf{S}ynthetic \textbf{T}ime--frequency condensation (\textbf{\fallnest{}}), which shrinks the training data rather than the model. It distills a large, imbalanced IMU dataset into a tiny, class-balanced synthetic set on which a lightweight CNN learns discriminative fall patterns. \fallnest{} builds on CondTSC~\cite{ref:condtsc_paper} and differs in four ways: (i) a fall-specific 7-channel input (tri-axial accelerometer, tri-axial gyroscope, accelerometer vector magnitude); (ii) class-balanced K-Means initialization, which balances the set by construction; (iii) normalized gradient matching with gradient clipping, which stabilizes optimization across datasets of very different sizes; and (iv) strict subject-wise splits with naturally imbalanced test sets. Raw signals are left unfiltered so that sharp impact peaks are preserved. On SisFall~\cite{ref:sisfall_dataset}, KFall~\cite{ref:KFall_dataset}, and MobiFall~\cite{ref:mobifall_dataset}, a 102\,K-parameter TinyCNN trained on 32 to 146 synthetic windows (about 1\% of the training windows) reaches 93.5 to 95.8\% mean accuracy and 0.89 to 0.98 PR-AUC over five seeds. The roughly 100-fold reduction applies to the downstream training set, not to the data accessed during condensation, which uses all of $\T$; a real validation set only selects the checkpoint.
\\
Our contributions are: a dual-domain condensation framework that turns imbalanced IMU recordings into compact, class-balanced training sets without re-sampling; a dual-objective scheme pairing normalized, clipped gradient matching with MMD alignment in both domains, plus multi-view Fourier augmentation; and a rigorous evaluation protocol (subject-wise splits, natural test priors, accuracy, Macro-F1, PR-AUC over five seeds) with ablations and a PCA analysis of real versus synthetic embeddings. All datasets are public, and code will be released upon acceptance.

\Section{Method}
\label{sec:method}

\begin{figure}[t]
 \includegraphics[width=\textwidth]{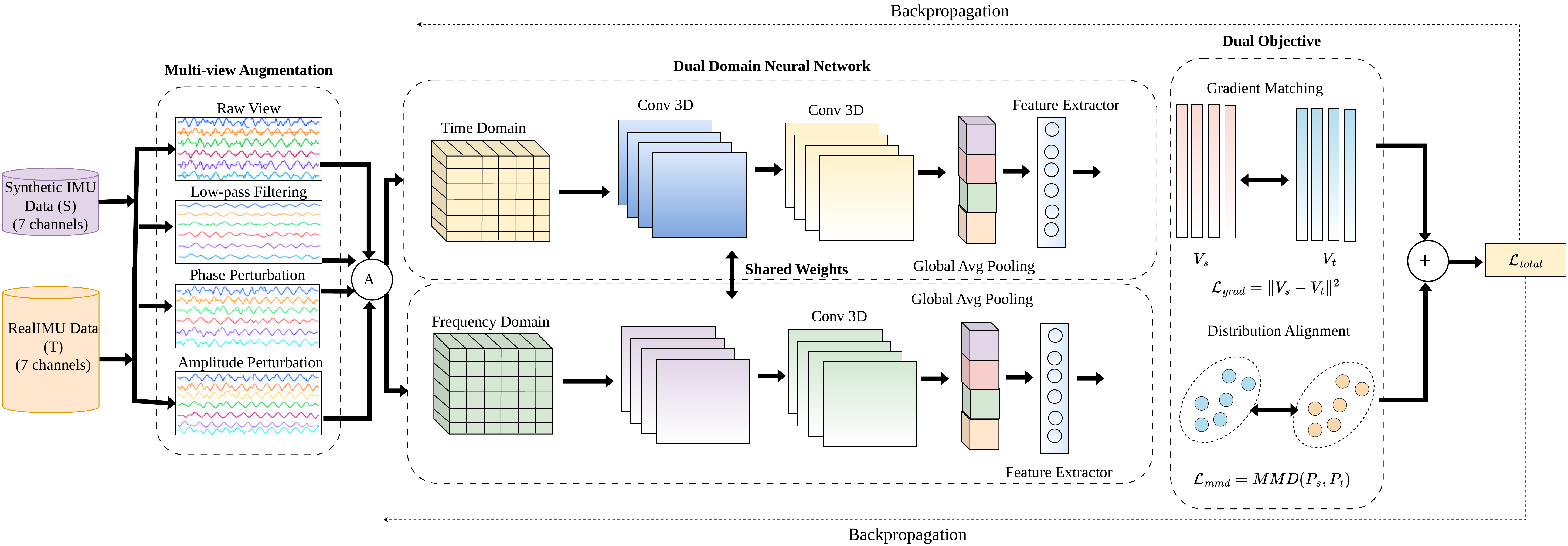}
  \caption{\fallnest{} pipeline. Imbalanced IMU windows $\T$ are condensed into a class-balanced set $\Sset$ via K-Means initialization, four-view Fourier augmentation, and joint gradient/MMD matching in the time and frequency domains. A TinyCNN is then trained on the obtained condensed set $\Sset$ and tested on $\Ttest$.}
  \label{fig:overview}
\end{figure}

\fallnest{} condenses a large, imbalanced IMU training set into a tiny, class-balanced synthetic set, and a lightweight classifier is trained on that set alone (Fig.~\ref{fig:overview}). Recordings become subject-disjoint windows; a balanced synthetic set is initialized with class-wise K-Means and refined in time and frequency domains through four augmented views by a joint normalized gradient-matching and MMD objective (Alg.~\ref{alg:fallnest}). A TinyCNN is then trained on the condensed set ($\Sset$), selected on a real validation set ($\V$), and evaluated on a subject-independent, naturally imbalanced test set ($\Ttest$).

\noindent
The real training set is $\T=\{(x_i,y_i)\}_{i=1}^{n}$ with $x_i\in\mathbb{R}^{C\times L}$ ($C=7$ channels, $L$ time steps) and $y_i\in\{0,1\}$ (0: ADL, 1: Fall); $\T_c$ denotes class $c$, and the validation set $\V$ and test set $\Ttest$ are subject-disjoint from $\T$ and each other. The synthetic set $\Sset$ holds $spc$ learnable tensors per class with fixed labels:
\begin{equation}
spc=\lceil 0.01\,n/2\rceil,\qquad |\Sset|=2\,spc\approx 0.01\,|\T|,
\label{eq:spc}
\end{equation}
with $n=|\T|$ counting \emph{all} training windows. With $\mathcal{L}_{\mathcal{D}}(\theta)$ the mean cross-entropy of a network with parameters $\theta$ on set $\mathcal{D}$, condensation seeks
\begin{equation}
\min_{\Sset}\;\mathcal{L}_{\T}\big(\theta^{*}(\Sset)\big)\quad\text{s.t.}\quad \theta^{*}(\Sset)=\arg\min_{\theta}\mathcal{L}_{\Sset}(\theta),
\label{eq:bilevel}
\end{equation}
i.e.\ a network trained only on $\Sset$ should perform on real data like one trained on $\T$. Solving Eq.~\eqref{eq:bilevel} exactly needs a training run per update, so we replace it with the matching objectives of Eqs.~\eqref{eq:grad} and \eqref{eq:emb}.

\SubSection{Data Preparation}
\label{sec:prep}
SisFall, KFall, and MobiFall contain 15, 15, and 4 fall types and 19, 21, and 9 ADL types; all falls map to $y{=}1$ and all ADLs to $y{=}0$. Recordings are resampled to 200\,Hz and segmented with a 2\,s stride into windows of $L=2000$, $1200$, and $1000$ samples (10, 6, and 5\,s, chosen from each dataset's mean fall duration). Each window inherits the binary label of its source trial, since only trial-level labels are available, and windows containing transitions are kept so that training matches deployment. Splits are \emph{by subject} (26/6/6, 16/4/4, and 15/4/13 subjects for SisFall, MobiFall, and KFall; all five seeds share the same split; see Table~\ref{tab:dataset_stats}). Besides the tri-axial accelerometer and gyroscope, each window carries the accelerometer vector magnitude $\mathrm{AVM}_t=\sqrt{a_{x,t}^{2}+a_{y,t}^{2}+a_{z,t}^{2}}$ of the \emph{raw} axes $a_{\cdot,t}$ at step $t$, an orientation-independent impact measure. Each of the 7 channels is standardized with the mean and std of $\T$. No global low-pass filter is applied, since it would smooth the impact peak; smoothing enters only as an augmentation view.

\SubSection{Class-Balanced K-Means Initialization}
\label{sec:init}
For each class $c$ we cluster the flattened windows $\mathrm{vec}(x)$ of $\T_c$ into $spc$ clusters (Eq.~\eqref{eq:spc}) with centroids $\mu_{c,k}$,
\begin{equation}
\{\mu_{c,k}\}_{k=1}^{spc}=\arg\min_{\{\mu_k\}}\sum_{x\in\T_c}\min_{k}\big\|\mathrm{vec}(x)-\mu_k\big\|_2^{2},
\label{eq:kmeans}
\end{equation}
and reshape each $\mu_{c,k}$ to $C\times L$ as an initial synthetic sample. Both classes get exactly $spc$ samples, so $\Sset$ is balanced whatever the real fall/ADL ratio, and centroids spread over the dominant modes of each class, which matters most for the heterogeneous ADL class. Centroids (Eq.~\eqref{eq:kmeans}) are averages and blur sharp peaks, so restoring impact structure is the job of the refinement.

\SubSection{Dual-Domain Representation}
\label{sec:dual}
The time-domain input is the window itself; the frequency-domain input is the channel-averaged magnitude spectrogram
\begin{equation}
\mathbf{P}(x)=\tfrac{1}{C}\textstyle\sum_{j=1}^{C}\big|\mathrm{STFT}(x_{j})\big|\in\mathbb{R}^{129\times F},
\label{eq:stft}
\end{equation}
where $x_j$ is channel $j$, $n_{\mathrm{fft}}=256$ (giving $n_{\mathrm{fft}}/2+1=129$ frequency bins), hop $64$, and $F=28$, $15$, $12$ frames for SisFall, KFall, MobiFall. Each domain $d\in\{\mathsf{t},\mathsf{f}\}$ has its own backbone $\phi_d$, shared between synthetic and real data.

\SubSection{Multi-View Fourier Augmentation}
\label{sec:aug}
Each sample $S$ is observed through four views $v$: the raw signal and three Fourier-domain views produced in sequence. The low-pass view $S_{\mathrm{LPF}}=\mathrm{FFT}^{-1}(H_{f_c}\odot\mathrm{FFT}(S))$ applies a hard mask $H_{f_c}$ ($\odot$: elementwise product) at cutoff $f_c=5$\,Hz and emphasizes macro-motion. The phase-perturbed view $S_{\mathrm{PP}}$ rebuilds $S_{\mathrm{LPF}}$ from its spectrum magnitude and its phase plus noise $\boldsymbol{\Phi}\sim\mathcal{N}(0,0.1^2)$ per frequency bin (tolerance to small timing shifts). The magnitude-perturbed view $S_{\mathrm{MP}}$ adds $\boldsymbol{\Gamma}\sim\mathcal{N}(0,0.1)$ to each magnitude bin of $S_{\mathrm{PP}}$ (intensity variation). All views are returned to the time domain. 

\SubSection{Dual-Objective Matching}
\label{sec:obj}
For domain $d$, view $v$, and class $c$, we draw random initial parameters $\theta_d^{0}$ of $\phi_d$, take $N$ differentiable SGD steps on $v(\Sset_c)$ giving $\theta_d^{\Sset}$ and $M$ SGD steps on real batches of $\T_c$ giving $\theta_d^{\T}$, and form the displacements $G^{\Sset}=\theta_d^{0}-\theta_d^{\Sset}$ and $G^{\T}=\theta_d^{0}-\theta_d^{\T}$. They are aligned with a normalized squared distance, which, unlike cosine matching, penalizes both direction and magnitude mismatch:
\begin{equation}
\mathcal{L}_{\mathrm{grad}}^{(v,d)}=\sum_{c}\frac{\|G^{\Sset}-G^{\T}\|_2^{2}}{\|G^{\Sset}\|_2^{2}+\|G^{\T}\|_2^{2}+\epsilon}.
\label{eq:grad}
\end{equation}
Here $\epsilon>0$ is a small constant, and each class-wise term lies in $[0,2]$ regardless of dataset size, comparable to the bounded MMD term. Penultimate-layer embeddings $z^{\Sset}$ and $z^{\T}$ of synthetic and real samples of the same class are aligned with the squared MMD under a Gaussian kernel of bandwidth $\sigma=1$:
\begin{equation}
\mathcal{L}_{\mathrm{emb}}^{(v,d)}=\sum_{c}\mathrm{MMD}^{2}\big(z^{\Sset},z^{\T}\big).
\label{eq:emb}
\end{equation}
Eq.~\eqref{eq:grad} aligns \emph{how the network learns}, Eq.~\eqref{eq:emb} \emph{what it represents}, and the total objective (weight $\lambda_{\mathrm{emb}}$ on the embedding term) is
\begin{equation}
\mathcal{L}=\sum_{v}\sum_{d}\Big(\mathcal{L}_{\mathrm{grad}}^{(v,d)}+\lambda_{\mathrm{emb}}\,\mathcal{L}_{\mathrm{emb}}^{(v,d)}\Big),\quad \lambda_{\mathrm{emb}}=1.
\label{eq:obj}
\end{equation}
We backpropagate through the unrolled inner loop, clip $\nabla_{\Sset}\mathcal{L}$ (Eq.~\eqref{eq:obj}) elementwise to $[-1,1]$, and update $\Sset$ with Adam (lr $=0.01$).

\begin{algorithm}[t]
\centering

\caption{\fallnest{} condensation and training}
\label{alg:fallnest}
\footnotesize
\begin{algorithmic}[1]
\Require Train set $\T$, $spc$, views (raw, LPF, PP, MP), domains $\{\mathsf{t},\mathsf{f}\}$, $N$, $M$, $\lambda_{\mathrm{emb}}$
\Ensure Trained classifier $f_{\theta^{*}}$
\State $\Sset_c\gets$ K-Means centroids of $\T_c$, $c\in\{0,1\}$ \Comment{Eq.~\eqref{eq:kmeans}}
\For{epoch $=1$ \textbf{to} 100}
  \For{each view $v$, domain $d$, class $c$}
    \State draw $\theta_d^{0}$; run $N$ SGD steps on $v(\Sset_c)$ and $M$ on batches of $\T_c$
    \State $\mathcal{L}\mathrel{+}=\mathcal{L}_{\mathrm{grad}}+\lambda_{\mathrm{emb}}\mathcal{L}_{\mathrm{emb}}$ \Comment{Eqs.~\eqref{eq:grad},~\eqref{eq:emb}}
  \EndFor
  \State $\Sset\gets\mathrm{Adam}\big(\Sset,\ \mathrm{clip}_{[-1,1]}(\nabla_{\Sset}\mathcal{L})\big)$; stop early if converged
\EndFor
\State Train a fresh TinyCNN on $\Sset$; keep the epoch with best Macro-F1 on $\V$
\end{algorithmic}
\end{algorithm}

\SubSection{Downstream Training, Backbone, and Evaluation}
\label{sec:downstream}
\label{sec:tinycnn}
\label{sec:eval}
A fresh TinyCNN is trained on $\Sset$ only (cross-entropy, 100 epochs, Adam lr $0.001$, batch size 16). Each backbone has three Conv1D--BatchNorm--ReLU blocks, adaptive average pooling, and a fully connected layer; the deployed classifier is the time-domain TinyCNN alone, and the frequency branch is used only during condensation. The model has 102\,K parameters ($\approx$400\,KB in float32) and needs about 103\,M MACs ($L{=}1000$) to 206\,M MACs ($L{=}2000$) per inference, with 2.15\,ms median CPU latency (Kaggle CPU, single-threaded PyTorch, batch size 1). These are architecture properties, not condensation effects; microcontroller validation is future work. We keep the checkpoint with the highest Macro-F1 on the real validation set, $e^{*}=\arg\max_{e}\mathrm{MacroF1}(f_{\theta_e};\V)$, where $f_{\theta_e}$ is the classifier after epoch $e$; since condensation uses $\T$ alone, $\V$ is the only other real data that enters. The checkpoint is evaluated once on $\Ttest$. Each of five seeds draws a new K-Means initialization, runs an independent condensation, and trains a new classifier on the same subject-wise split. We report mean $\pm$ std of Accuracy, Macro-F1, and PR-AUC (Fall positive), plus fall precision, recall, specificity, and false alarms per hour (FA/h, false-positive windows per ADL hour at the 2\,s stride; a window-level estimate on laboratory data). Overlapping windows from one trial are correlated, so window counts overstate independent events.

\SubSection{Theoretical Analysis of the Objective}
\label{sec:analysis}
Fix a domain, view, and class, and let $\Delta=\theta^{\Sset}-\theta^{\T}=G^{\T}-G^{\Sset}$ be the parameter gap, $r$ the class-wise summand of Eq.~\eqref{eq:grad}, and $P$ the real-batch distribution. Assume the real loss $L_P$ (expected cross-entropy under $P$) is $\Lambda$-Lipschitz and $\beta$-smooth on a convex set containing the iterates, SGD steps (rate $\eta$, gradient norm $\le g_{\max}$) are bounded by $\eta g_{\max}$ (so $\Omega^2=\eta^2g_{\max}^2(N^2+M^2)+\epsilon$), the head loss is approximable in the RKHS of the Gaussian kernel (norm $\le B$, error $\le\kappa$), and the $n_T$ real embeddings are i.i.d.\ from $P$ given $\Sset$.
\begin{corollary}
\label{cor:main}
Let $\delta\in[r_0,1)$ with $r_0$ the value of $r$ at $\epsilon=0$, $s=1/(1-\delta)$, $\rho_\pm=s\pm\sqrt{s^2-1}$, and $K(\delta)=1+\rho_+^2$. Then $\cos\angle(G^{\Sset},G^{\T})\ge1-\delta$, $\|G^{\Sset}\|/\|G^{\T}\|\in[\rho_-,\rho_+]$, $|L_P(\theta^{\Sset})-L_P(\theta^{\T})|\le\Lambda\|\Delta\|$, and, with probability at least $1-\eta'$ over the real batch,
\begin{align}
\|\Delta\|&\le\min\Big\{\Omega\sqrt{r},\ \sqrt{\delta\,K(\delta)}\,\|G^{\T}\|\Big\},
\label{eq:cor_a}\\
L_P(\theta^{\Sset})&\le L_{\Sset}(\theta^{\Sset})+B\Big[\sqrt{\mathcal{L}_{\mathrm{emb}}}+\tfrac{1+\sqrt{\ln(1/\eta')}}{\sqrt{n_T}}\Big]+2\kappa.
\label{eq:cor_c}
\end{align}
\end{corollary}
\noindent Eq.~\eqref{eq:cor_a} shows that Eq.~\eqref{eq:grad} controls both direction and scale of the update, which bounds the real-loss gap to the real-trained network, while Eq.~\eqref{eq:cor_c} lets the MMD term of Eq.~\eqref{eq:emb} bound the real loss by the synthetic loss. The constants are unmeasured, so the bounds are qualitative and concern the training distribution, not unseen subjects (assessed empirically below).
\Section{Experimental Setup}
\label{sec:experimental_setup}
\SubSection{Datasets and Implementation}
\label{subsec:datasets}
We evaluate on three binary fall-detection datasets: SisFall (38 subjects), KFall (32), and MobiFall (24). Dataset statistics and subject-wise splits are summarized in Table~\ref{tab:dataset_stats}. 
\begin{wraptable}{r}{0.45\textwidth}
\caption{\label{tab:dataset_stats}%
    Dataset statistics after subject-wise splitting (2\,s stride). $\T$ is the only set used for condensation, $\V$ only selects the checkpoint, and $\Ttest$ is used once and keeps its natural class distribution.}
\renewcommand{\baselinestretch}{1}\footnotesize
\resizebox{\linewidth}{!}{%
\begin{tabular}{l c c c c c c}
        \toprule
        \textbf{Dataset} & \textbf{Split} & \textbf{Subj.} & \textbf{Win.} & \textbf{Total} & \textbf{ADL} & \textbf{Fall} \\
        \midrule
        \multirow{3}{*}{SisFall}  & $\T$        & 26 & 10\,s & 14,528 & 10,259 & 4,269 \\
                                   & $\V$        & 6  &       &  2,962 &  2,287 &   675 \\
                                   & $\Ttest$    & 6  &       &  2,778 &  2,328 &   450 \\
        \midrule
        \multirow{3}{*}{MobiFall} & $\T$        & 16 & 5\,s  &  3,146 &  2,568 &   578 \\
                                   & $\V$        & 4  &       &  1,000 &    856 &   144 \\
                                   & $\Ttest$    & 4  &       &    571 &    427 &   144 \\
        \midrule
        \multirow{3}{*}{KFall}    & $\T$        & 15 & 6\,s  &  3,574 &  2,272 & 1,302 \\
                                   & $\V$        & 4  &       &  1,046 &    583 &   463 \\
                                   & $\Ttest$    & 13 &       &  2,933 &  1,922 & 1,011 \\
        \bottomrule
    \end{tabular}}
\end{wraptable}
 All condensation hyperparameters are fixed across datasets, except the real-data batch size, with $spc=\{73,16,18\}$ for SisFall, MobiFall, and KFall, respectively. Condensation uses at most 100 epochs, learning rate $0.01$, $\lambda_{\mathrm{emb}}=1$, $N=15$, $M=100$, inner learning rate $0.001$, patience 10, and $f_c=5$\,Hz. The real batch size is 128 for SisFall and 64 for KFall and MobiFall. No dataset-specific hyperparameter tuning is performed. All models are implemented in PyTorch and trained on a 16 GB P100 GPU. Experiments use five fixed random seeds; seed 42 is used for representative single-run results. Since the subject splits remain fixed across seeds, the reported standard deviations capture initialization variability rather than variability across subject partitions. 
\noindent

\Section{Results and Discussion}
\label{sec:results}

\SubSection{Classification Performance}
\label{subsec:classification}

Table~\ref{tab:main_results} reports test Accuracy, Macro-F1, and PR-AUC. Using only 1\% of training windows, \fallnest{} achieves 93.5--95.8\% accuracy and 0.891--0.975 PR-AUC, exceeding the always-ADL accuracy baselines by 9.7, 20.7, and 30.3 percentage points on SisFall, MobiFall, and KFall, respectively. Compared with full-data training, accuracy drops by 6.4, 4.5, and 3.6 points, respectively; SisFall also shows the largest PR-AUC reduction (0.999 to 0.891). 
Although five-seed variability is modest (1.0--1.7 accuracy points; 0.013--0.023 PR-AUC), fixed test subjects mean these standard deviations reflect initialization variability only. SisFall's larger performance gap may relate to its lower fall prevalence, longer windows that dilute impact signatures (Table~\ref{tab:dataset_stats}), and greater ADL heterogeneity relative to its 146-sample condensed set;these hypotheses require targeted evaluation. Table~\ref{tab:operating} presents the performance metrics in terms of  precision, recall, and specificity for each dataset.

\begin{table}[t]
    \centering
    \footnotesize
    \setlength{\tabcolsep}{5pt}
    \caption{\label{tab:main_results}%
    Main results on subject-independent, naturally imbalanced $\Ttest$. Rows $\Sset$ train on the condensed set (1\% of $\T$), mean $\pm$ std over 5 seeds; rows $\T$ train the same TinyCNN on the full set (single run). Size is the number of training windows. Always-ADL accuracy floors: 83.8/74.8/65.5\%; random-scorer PR-AUC floors: 0.162/0.252/0.345 (SisFall/MobiFall/KFall).}
    
    \resizebox{0.7\textwidth}{!}{%
    \begin{tabular}{l l r c c c}
        \toprule
        \textbf{Dataset} & \textbf{Train set} & \textbf{Size} &
        \textbf{Acc (\%)} & \textbf{Macro-F1} & \textbf{PR-AUC} \\
        \midrule
        
        \multirow{2}{*}{SisFall}
          & $\Sset$ (condensed) & 146
          & $93.50\!\pm\!1.02$
          & $0.887\!\pm\!0.016$
          & $0.891\!\pm\!0.023$ \\
          
          & $\T$ (full) & 14,528
          & 99.85
          & 0.990
          & 0.999 \\
          
        \midrule
        
        \multirow{2}{*}{MobiFall}
          & $\Sset$ (condensed) & 32
          & $95.48\!\pm\!1.73$
          & $0.942\!\pm\!0.023$
          & $0.975\!\pm\!0.017$ \\
          
          & $\T$ (full) & 3,146
          & 100.00
          & 1.000
          & 1.000 \\
          
        \midrule
        
        \multirow{2}{*}{KFall}
          & $\Sset$ (condensed) & 36
          & $95.80\!\pm\!1.68$
          & $0.954\!\pm\!0.018$
          & $0.973\!\pm\!0.013$ \\
          
          & $\T$ (full) & 3,574
          & 99.39
          & 0.999
          & 1.000 \\
          
        \bottomrule
    \end{tabular}%
    }
\end{table}


\SubSection{Comparison with Equal-Budget Real Subsets}
\label{subsec:coreset}
Does condensation beat simply training on a 1\% real subset? Table~\ref{tab:coreset} compares \fallnest{} with random real subsets of the same size $|\Sset|$, drawn either uniformly or with equal numbers per class, under the same TinyCNN, checkpoint selection, and five-seed protocol.

\begin{table}[t]
    \centering
    \scriptsize

    \begin{minipage}[t]{0.6\columnwidth}
        \centering
        \setlength{\tabcolsep}{2.5pt}
        \caption{Equal-budget comparison, Acc / Macro-F1 / PR-AUC
        (mean over 5 seeds). Random subsets contain $|\Sset|$ windows
        drawn from $\T$.}
        \label{tab:coreset}

        \resizebox{\linewidth}{!}{%
        \begin{tabular}{l c c c}
            \toprule
            \textbf{Training set} &
            \textbf{SisFall} &
            \textbf{MobiFall} &
            \textbf{KFall} \\
            \midrule

            Real (uniform)
                & 94.0/0.90/0.88
                & 91.73/0.892/0.915
                & 89.74/0.89/0.90 \\

            Real (balanced)
                & 92.55/0.879/0.765
                & 93.06/0.916/0.932
                & 95.2/0.95/0.86 \\

            \fallnest{} (ours)
                & 93.50/0.887/0.891
                & 95.48/0.942/0.975
                & 95.80/0.954/0.973 \\

            \bottomrule
        \end{tabular}%
        }
    \end{minipage}
    \hfill
    \begin{minipage}[t]{0.37\columnwidth}
        \centering
        \setlength{\tabcolsep}{2pt}
        \caption{Operating-point metrics (seed 42)}
        \label{tab:operating}

        \resizebox{\linewidth}{!}{%
        \begin{tabular}{l c c c}
            \toprule
            \textbf{Dataset} &
            \textbf{Precision} &
            \textbf{Recall} &
            \textbf{Specificity} \\
            \midrule

            SisFall
                & 0.84 & 0.86 & 0.97 \\

            MobiFall
                & 0.90 & 0.97 & 0.96 \\

            KFall
                & 0.96 & 0.94 & 0.98 \\

            \bottomrule
        \end{tabular}%
        }
    \end{minipage}
\end{table}
On MobiFall and KFall, where only 32 and 36 samples are available, \fallnest{} leads in accuracy and especially PR-AUC (0.975 and 0.973 versus at most 0.932 and 0.90 for the random subsets), so condensation helps most where random draws leave ADL regions uncovered, whereas synthetic samples span both classes (Fig.~\ref{fig:pca_embeddings}). On SisFall the uniform subset is competitive in accuracy (94.0 versus 93.5) and Macro-F1, so there condensation mainly improves ranking quality over the class-balanced subset (PR-AUC 0.891 versus 0.765), and its benefit over uniform sampling is not established. Balancing alone does not explain the gains on the two smaller datasets, since the class-balanced subset trails \fallnest{} in all three metrics there.
\SubSection{Comparison with Full-Data Methods}
\label{subsec:sota}
Table~\ref{tab:sota} lists recent full-data methods, which report 98--99.8\% accuracy on these datasets, with the \fallnest{} rows for the same datasets below. Each published row gives one metric only, accuracy for most methods and PR-AUC for the BiLSTM--BiGRU model, so a full comparison on both metrics is not possible. Against these methods, \fallnest{} gives up about 3--6 points of accuracy (for example 99.3\% versus 93.5\% on SisFall) while using about two orders of magnitude fewer training windows. The comparison is not like for like, because published protocols differ in splitting, class balance, and metrics. LiteFallNet~\cite{owusu2025litefallnet} and BiLSTM--BiGRU~\cite{ahirwar2025optimized} use random window-level splits, which can inflate results through temporal leakage, whereas \cite{zhang2024effective} and \cite{yu2020novel} use variants of subject-wise cross-validation. One could suspect that published numbers are inflated by window-level splits, but our own full-data TinyCNN under strict subject-wise splitting reaches 99.85\%, 100.00\%, and 99.39\%, at the level of the published methods, so leakage is not the main explanation. These benchmarks appear saturated under either protocol, and the 3--6 point gap is mostly the price of the 100-fold data reduction. We do not claim parity with full-data training. The case for condensation is use cases in which storing or processing the full corpus is impractical, such as rapid prototyping or on-device personalization.


\begin{figure*}[t]
    \centering
    \begin{minipage}[c]{0.43\textwidth}
        \centering
        \includegraphics[width=0.8\linewidth]{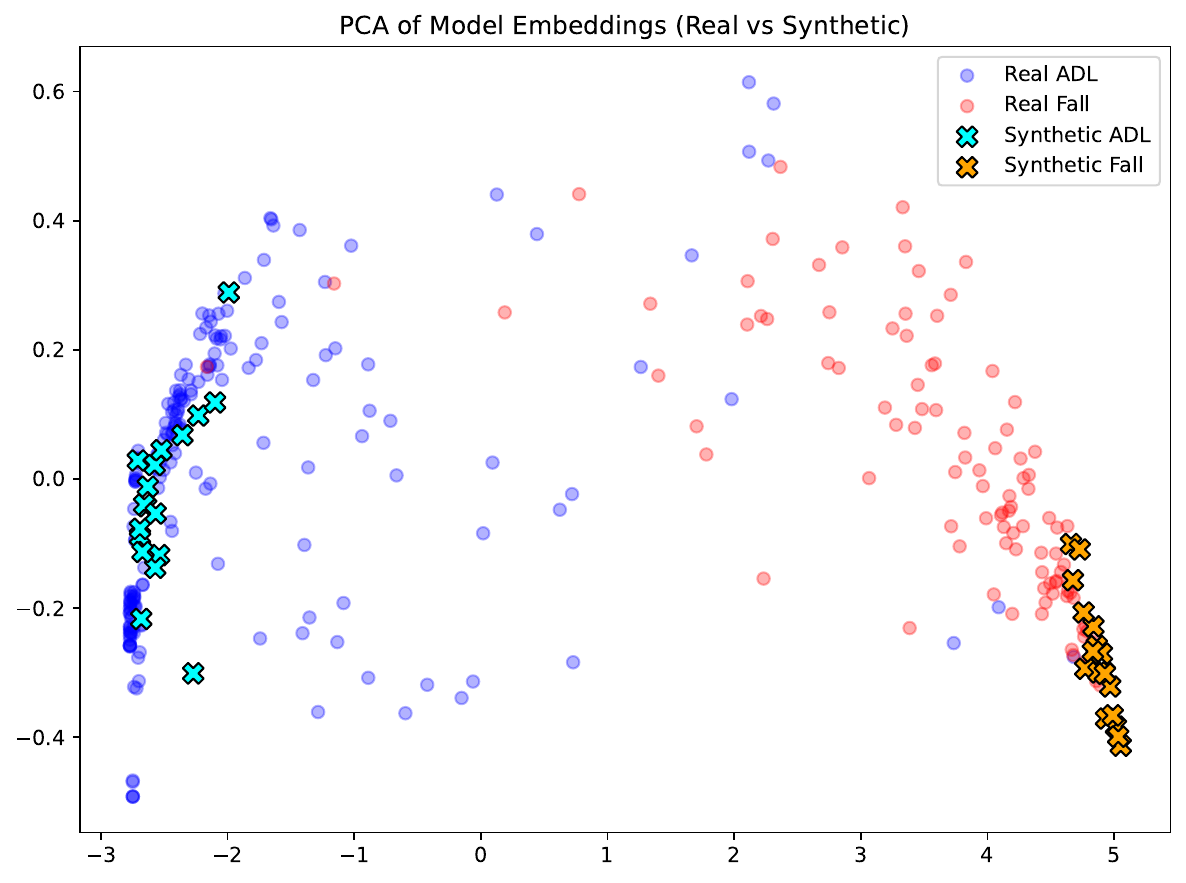}
        \caption{PCA projection of feature embeddings. Synthetic
        samples (crosses) lie within the real ADL and Fall distributions
        (circles).}
        \label{fig:pca_embeddings}
    \end{minipage}%
    \hfill
    \begin{minipage}[c]{0.5\textwidth}
        \vspace{-2em}\makeatletter\def\@captype{table}\makeatother  
        \centering
        \footnotesize
        \setlength{\tabcolsep}{5pt}
        \caption{%
        Published full-data methods (100\% of the training data) versus
        \fallnest{} (about 1\%). Splits and metrics differ, so the table is
        indicative only.}\label{tab:sota}
        \resizebox{\linewidth}{!}{%
        \begin{tabular}{l l l}
            \toprule
            \textbf{Method (year)} & \textbf{Dataset} & \textbf{Acc / PR-AUC} \\
            \midrule
            Deep learning (2024)~\cite{zhang2024effective}
                & SisFall / MobiFall & Acc $\approx$ 99.3 / 99.6\% \\
            LiteFallNet (2025)~\cite{owusu2025litefallnet}
                & KFall / SisFall & Acc $\approx$ 99.0 / 98--99\% \\
            BiLSTM--BiGRU (2025)~\cite{ahirwar2025optimized}
                & SisFall & PR-AUC $\approx$ 0.99 \\
            Hybrid KNN (2020)~\cite{yu2020novel}
                & KFall & Acc $\approx$ 99.8\% \\
            \midrule
            \textbf{\fallnest{} (1\% data)} & SisFall  & 93.5\% / 0.891 \\
            \textbf{\fallnest{} (1\% data)} & MobiFall & 95.5\% / 0.975 \\
            \textbf{\fallnest{} (1\% data)} & KFall    & 95.8\% / 0.973 \\
            \bottomrule
        \end{tabular}%
        }
    \end{minipage}
\end{figure*}
  

\SubSection{Ablation Studies}
\label{subsec:ablation}

Table~\ref{tab:ablation} reports single runs (seed 42), for different ablations. MobiFall is generally the most sensitive. Accuracy and PR-AUC sometimes diverge: on MobiFall, several variants lose 6--20 accuracy points while keeping PR-AUC at 0.97--0.98, suggesting an operating-point shift between validation (14.4\% fall prevalence) and test data (25.2\%). Frequency-only is the exception, with PR-AUC down to 0.71.

\begin{wraptable}{r}{0.5\textwidth}\vspace{-1.5em}
\centering\scriptsize\setlength{\tabcolsep}{3pt}
\caption{\label{tab:ablation}Ablations, Acc (\%) / PR-AUC, single run (seed 42). Full = dual domain, K-Means init., multi-view aug., combined loss.}
\resizebox{\linewidth}{!}{%
\begin{tabular}{l c c c}
\toprule
\textbf{Config.} & \textbf{SisFall} & \textbf{MobiFall} & \textbf{KFall} \\
\midrule
\textbf{Full} & \textbf{95.03/0.90} & \textbf{98.07/0.996} & \textbf{96.59/0.99} \\
\hline
Freq. only & 90.78/0.896 & 70.75/0.71 & 91.9/0.93 \\
Time only & 91.47/0.88 & 97.20/0.995 & 92.57/0.91 \\
\hline
Random init & 93.30/0.79 & 92.47/0.98 & 96.9/0.95 \\
\hline
No aug. & 91.68/0.83 & 78.11/0.978 & 95.02/0.95 \\
\hline
Grad. only & 93.74/0.85 & 84.41/0.97 & 94.6/0.93 \\
MMD only & 88.66/0.84 & 91.07/0.97 & 93.7/0.92 \\
\bottomrule
\end{tabular}}
\vspace{-0.8em}
\end{wraptable}
\noindent
\emph{Domain.} Frequency-only loses 4.3, 27.3, and 4.7 accuracy points on SisFall, MobiFall, and KFall, and time-only loses 3.6, 0.9, and 4.0. The dual-domain gain is clear on SisFall and KFall, while the MobiFall time-only result is within seed noise. The frequency-only collapse on MobiFall may reflect its 12 spectrogram frames.
\\
\emph{Initialization.} Random initialization costs 1.7 and 5.6 points on SisFall and MobiFall and gains 0.3 on KFall; on SisFall, PR-AUC drops from 0.90 to 0.79.
\\
\noindent
\emph{Augmentation.} Removing it costs 3.4, 20.0, and 1.6 points, most on MobiFall, where only 32 samples are condensed. Its preserved PR-AUC suggests an operating-point effect.
\\
\emph{Loss.} Either loss alone is below the combination, except gradient-only on SisFall (within noise). Gradient-only beats MMD-only on SisFall and KFall, and the reverse holds on MobiFall. The ablations do not isolate AVM, normalized versus cosine matching, or clipping, and we include no CondTSC or other coreset/condensation baselines (e.g.\ herding, k-center, DM), so these differences from CondTSC are design choices, not validated gains.
\SubSection{Training and Inference Cost}
\label{subsec:efficiency}

Condensation reduces the \emph{training data} and offline compute, not inference latency, memory, or energy, so we report both and claim no deployment speedup from it. Table~\ref{tab:efficiency} trains architectures from 0.8\,K to about 1\,M parameters on the same condensed sets.

\begin{wraptable}{r}{0.58\textwidth}
  \vspace{-1em}
  \centering
  \scriptsize
  \setlength{\tabcolsep}{3pt}
  \caption{\label{tab:efficiency}%
  Model-scale sweep on the same condensed $\Sset$ (Acc / Macro-F1; Acc only for KFall; seed 42) and cost of the TinyCNN used throughout.}
  \resizebox{\linewidth}{!}{%
  \begin{tabular}{l c c c c}
    \toprule
    \textbf{Arch.} & \textbf{Params} & \textbf{SisFall} & \textbf{MobiFall} & \textbf{KFall} \\
    \midrule
    MicroCNN  & 0.8K  & 92.3/0.87 & 68.1/0.67  & 73.6 \\
    SmallCNN  & 10K   & 93.3/0.88 & 76.7/0.75  & ---  \\
    \textbf{TinyCNN} & \textbf{102K} & \textbf{95.0/0.91} & \textbf{98.1/0.975} & \textbf{96.6} \\
    LargeMLP  & $\sim$500K--1M & 93.3/0.88 & 75.1/0.73 & 96.8 \\
    \midrule
    \multicolumn{2}{l}{MACs / size / CPU latency} & \multicolumn{3}{c}{103--206\,M / $\approx$400\,KB / 2.15\,ms} \\
    \multicolumn{2}{l}{Condensation time}    & $\sim$82\,min & $\sim$21\,min & $\sim$26\,min \\
    \multicolumn{2}{l}{Train: $\Sset$ vs.\ full $\T$} & $\sim$8 vs 37\,min & $\sim$3 vs 18\,min & $\sim$4.5 vs 25\,min \\
    \bottomrule
  \end{tabular}}
  \vspace{-0.8em}
\end{wraptable}
\noindent
TinyCNN is best on SisFall and MobiFall and within 0.2 points on KFall (96.6 versus 96.8 for LargeMLP). Smaller CNNs underfit unevenly: MicroCNN is 2.7 points below TinyCNN on SisFall but 30.0 on MobiFall, below the 74.8\% always-ADL floor, while SmallCNN (76.7\%) and LargeMLP (75.1\%) barely exceed it. On SisFall the high floor (83.8\%) hides this, and Macro-F1 (0.87 versus 0.91) is the better indicator. LargeMLP differs in architecture as well as size, so the sweep does not isolate parameter count. TinyCNN needs 103--206\,M MACs, about 400\,KB, and 2.15\,ms median CPU latency (single-threaded PyTorch, batch size 1); we did not test it on a microcontroller. Training on $\Sset$ is 4.6 to 6 times faster than on the full $\T$, but condensation plus training (about 90, 24, and 30.5\,min) exceeds one full-data run (37, 18, and 25\,min), so condensation pays off only when many architectures or hyperparameters are tried on one condensed set.
\Section{Conclusion}
\label{sec:conclusion}
We proposed \textbf{Fall} detection framework via \textbf{N}ormalized-gradient and \textbf{E}mbedding-matched \textbf{S}ynthetic \textbf{T}ime--frequency condensation (\textbf{\fallnest{}}), a dual-domain dataset condensation framework for wearable fall detection that distills a large, imbalanced IMU training set into a tiny, class-balanced synthetic set using class-balanced K-Means initialization, a 7-channel input with AVM, normalized and clipped gradient matching with MMD alignment, and multi-view Fourier augmentation. Under strict subject-wise splits and naturally imbalanced test sets, a 102\,K-parameter TinyCNN trained on 32 to 146 synthetic windows (about 1\% of the training windows) reaches 93.5--95.8\% mean accuracy and 0.89--0.98 PR-AUC over five seeds, 3.6 to 6.4 points below full-data training. The ablations indicate that K-Means initialization, multi-view augmentation, and the combined loss each matter, most clearly where the synthetic set is smallest. Inference cost was measured on a CPU only; future work includes false-alarm analysis on free-living data and microcontroller deployment.\newpage
\Section{References}
\label{sec:refs}
 
\bibliographystyle{IEEEbib}
\bibliography{refs}

@misc{ref:who_falls,
  author       = {World Health Organization},
  title        = {Falls},
  year         = {2024},
  url          = {https://www.who.int/news-room/fact-sheets/detail/falls},
  note         = {Accessed: 2024-09-15}
}

@article{ref:wearable_review,
  author={Singh, Avneesh and Rehman, Saqib Ur and Yongchareon, Somsak and Chong, Peter H. J.},
  journal={IEEE Sensors Journal},
  title={Sensor Technologies for Fall Detection Systems: A Review},
  year={2020},
  volume={20},
  number={13},
  pages={6889-6919},
  doi={10.1109/JSEN.2020.2974950}
}

@article{ref:imbalance_challenge,
  author={Pannurat, Nannapas and Thiemjarus, Treethep and Nantajeewarawat, Ekarat},
  journal={Sensors},
  title={Automatic Fall Monitoring: A Review},
  year={2014},
  volume={14},
  number={7},
  pages={12900-12936},
  doi={10.3390/s140712900}
}

@article{ref:embedded_ml_constraints,
  author={Banbury, Christian and Hinkle, Jessica and Hu, Bo and Lu, Shuo and Masi, Jonathan and Melnyk, Orest and Miller, Alex and Ni, Yang and Saletore, Vikas and Smith, Jack and Trask, Andrew and Venkatraman, Shiva and Xu, Haoran and Yang, Bin and Young, David and Zhao, Haotian},
  title={The Challenges and Opportunities of Machine Learning for Embedded Systems},
  journal={Proc. 11th USENIX Conf. on Networked Systems Design and Implementation (NSDI)},
  year={2014},
  pages={317--329}
}

@article{ref:campanella2024,
  author={Campanella, Sara and Alnasef, Alaa and Falaschetti, Laura and Belli, Alberto and Pierleoni, Paola and Palma, Lorenzo},
  journal={IEEE Sensors Journal},
  title={A Novel Embedded Deep Learning Wearable Sensor for Fall Detection},
  year={2024},
  volume={24},
  number={9},
  pages={15219-15229},
  doi={10.1109/JSEN.2024.3375603}
}

@article{ref:dc_intro,
  author={Wang, Tongzhou and Zhu, Jun-Yan and Torralba, Antonio and Efros, Alexei A.},
  title={Dataset Distillation},
  journal={arXiv preprint arXiv:1811.10959},
  year={2020}
}

@inproceedings{ref:condtsc_paper,
  author    = {Liu, Zhanyu and Hao, Ke and Zheng, Guanjie and Yu, Yanwei},
  title     = {Dataset Condensation for Time Series Classification via Dual Domain Matching},
  booktitle = {Proceedings of the 30th ACM SIGKDD Conference on Knowledge Discovery and Data Mining},
  year      = {2024},
  pages     = {1980--1989},
  doi       = {10.1145/3637528.3671675},
  publisher = {ACM}
}

@article{ref:sisfall_dataset,
title={SisFall: A fall and movement dataset},
author={Sucerquia, Angela and L{'o}pez, Jos{'e} David and Vargas-Bonilla, Jes{'u}s Francisco},
journal={Sensors},
volume={17},
number={1},
pages={198},
year={2017},
publisher={MDPI}
}

@article{ref:KFall_dataset,
title={A large-scale open motion dataset (KFall) and benchmark algorithms for detecting pre-impact fall of the elderly using wearable inertial sensors},
author={Yu, Xiaoqun and Jang, Jaehyuk and Xiong, Shuping},
journal={Frontiers in Aging Neuroscience},
volume={13},
pages={692865},
year={2021},
publisher={Frontiers Media SA}
}

@inproceedings{ref:mobifall_dataset,
title={The MobiFall dataset: An initial evaluation of fall detection algorithms using smartphones},
author={Vavoulas, George and Pediaditis, Matthew and Spanakis, Emmanouil G and Tsiknakis, Manolis},
booktitle={13th IEEE International Conference on BioInformatics and BioEngineering},
pages={1--4},
year={2013},
organization={IEEE}
}

@article{zhang2024effective,
  title={An effective deep learning framework for fall detection: model development and study design},
  author={Zhang, Jinxi and Li, Zhen and Liu, Yu and Li, Jian and Qiu, Hualong and Li, Mohan and Hou, Guohui and Zhou, Zhixiong},
  journal={Journal of medical internet research},
  volume={26},
  pages={e56750},
  year={2024},
  publisher={JMIR Publications Toronto, Canada}
}

@article{owusu2025litefallnet,
  title={LiteFallNet: A lightweight deep learning model for efficient real-time fall detection},
  author={Owusu, Emmanuel and Acquah, Isaac and Asare, Michael Asiedu and Yeboah, Benjamin Appiah},
  journal={Digital Health},
  volume={11},
  pages={20552076251386698},
  year={2025},
  publisher={SAGE Publications Sage UK: London, England}
}

@article{ahirwar2025optimized,
  title={Optimized fall detection using hybrid BiLSTM BiGRU additive attention model and BAOA driven feature selection system},
  author={Ahirwar, Mithun Singh and Soni, Vaibhav},
  journal={Scientific Reports},
  volume={15},
  number={1},
  pages={39254},
  year={2025},
  publisher={Nature Publishing Group UK London}
}

@article{yu2020novel,
  title={A novel hybrid deep neural network to predict pre-impact fall for older people based on wearable inertial sensors},
  author={Yu, Xiaoqun and Qiu, Hai and Xiong, Shuping},
  journal={Frontiers in bioengineering and biotechnology},
  volume={8},
  pages={63},
  year={2020},
  publisher={Frontiers Media SA}
}
 
\end{document}


\title
{\large
\textbf{\fallnest{}: Dual-Domain Dataset Condensation of Imbalanced IMU Data for Wearable Fall Detection}
}

\author{%
Rashi Niyas P$^{\ast}$, Dr. Ritika Jain$^{\dag}$, and Dr. Hitika Tiwari$^{\ast}$\\[0.5em]
{\small\begin{minipage}{\linewidth}\begin{center}
\begin{tabular}{ccc}
$^{\ast}$Indian Institute of Technology Madras & \hspace*{0.5in} & $^{\dag}$Indian Institute of Technology Guwahati \\
Zanzibar && Assam, India \\
zda24m005@iitmz.ac.in, hitika@iitmz.ac.in && rjain1992@iitg.ac.in
\end{tabular}
\end{center}\end{minipage}}
}

\maketitle
\thispagestyle{empty}

\Section{Theoretical Analysis of the Objective}
\label{sec:analysis}
Fix a domain, view, and class, and let $\Delta=\theta^{\Sset}-\theta^{\T}=G^{\T}-G^{\Sset}$, $r$ the class-wise summand of Eq.~\eqref{eq:grad}, and $P$ the real-batch distribution. Assume $L_P$ is $\Lambda$-Lipschitz and $\beta$-smooth on a convex set containing the iterates, SGD steps are bounded by $\eta g_{\max}$ (giving $\Omega^2=\eta^2g_{\max}^2(N^2+M^2)+\epsilon$), the head loss is approximable in the RKHS of the Gaussian kernel (norm $\le B$, error $\le\kappa$), and real embeddings are i.i.d.\ from $P$ given $\Sset$.
\begin{corollary}
\label{cor:main}
Let $\delta\in[r_0,1)$ with $r_0$ the value of $r$ at $\epsilon=0$, $s=1/(1-\delta)$, $\rho_\pm=s\pm\sqrt{s^2-1}$, $C(\delta)=1+\rho_+^2$. Then $\cos\angle(G^{\Sset},G^{\T})\ge1-\delta$, $\|G^{\Sset}\|/\|G^{\T}\|\in[\rho_-,\rho_+]$, and
\begin{equation}
\|\Delta\|\le\min\Big\{\Omega\sqrt{r},\ \sqrt{\delta\,C(\delta)}\,\|G^{\T}\|\Big\}.
\label{eq:cor_a}
\end{equation}
Moreover $|L_P(\theta^{\Sset})-L_P(\theta^{\T})|\le\Lambda\|\Delta\|$, and with probability at least $1-\eta'$ over the real batch,
\begin{equation}
L_P(\theta^{\Sset})\le L_{\Sset}(\theta^{\Sset})+B\Big[\sqrt{\mathcal{L}_{\mathrm{emb}}}+\tfrac{1+\sqrt{\ln(1/\eta')}}{\sqrt{n_T}}\Big]+2\kappa.
\label{eq:cor_c}
\end{equation}
\end{corollary}
\subsection{Theoretical Analysis of the Objective}
\label{sec:analysis}
We ask what a small value of Eq.~\eqref{eq:obj} certifies about the bilevel goal of Eq.~\eqref{eq:bilevel}. Fix a domain $d$, view $v$, and class $c$. Write $G^{\Sset}=\theta^{0}-\theta^{\Sset}$ and $G^{\T}=\theta^{0}-\theta^{\T}$ (Eq.~\eqref{eq:disp}), where $\theta^{0}$ is the random initial parameter vector and $\theta^{\Sset},\theta^{\T}$ are the parameters after the synthetic and real inner loops, let $\Delta=\theta^{\Sset}-\theta^{\T}=G^{\T}-G^{\Sset}$ be their gap, and let $P$ be the distribution of the real batches used at this step (the empirical distribution of $\T_c$, passed through view $v$ if views are applied to real data), with $L_P(\theta)=\mathbb{E}_{x\sim P}\,\ell(f_\theta(x),c)$ the expected loss under $P$. We define $r$ and $\tau$ in Eq.~\eqref{eq:r_tau},
\begin{equation}
r=\frac{\|G^{\Sset}-G^{\T}\|^{2}}{\|G^{\Sset}\|^{2}+\|G^{\T}\|^{2}+\epsilon},\qquad \tau=\frac{\epsilon}{\|G^{\Sset}\|^{2}+\|G^{\T}\|^{2}},
\label{eq:r_tau}
\end{equation}
where $r$ is the class-wise summand and $\tau$ is the ratio of $\epsilon$ to the squared displacement norms, and we set $r_0=(1+\tau)\,r$, the value of $r$ with $\epsilon=0$. We assume $G^{\Sset},G^{\T}\neq0$.
 
We use four assumptions. For regularity, $L_P$ is $\Lambda$-Lipschitz and $\beta$-smooth on a convex set $\Theta$ containing $\theta^{0}$, $\theta^{\Sset}$, and $\theta^{\T}$, that is, $\|\nabla L_P(\theta)-\nabla L_P(\theta')\|\le\beta\|\theta-\theta'\|$. For bounded steps, every SGD step has learning rate $\eta$ and gradient norm at most $g_{\max}$, so each step has size at most $\eta g_{\max}$, hence $\|G^{\Sset}\|\le N\eta g_{\max}$ and $\|G^{\T}\|\le M\eta g_{\max}$, and we set $\Omega^{2}=\eta^{2}g_{\max}^{2}(N^{2}+M^{2})+\epsilon$. For kernel-learnable head loss, the embeddings $z=\phi_\theta(x)$, where $\phi_\theta$ is the embedding function of $f_\theta$, lie in a compact set $\mathcal{Z}$ and $H$ is the RKHS of the Gaussian kernel $k$ on $\mathcal{Z}$; for every $\theta\in\Theta$ the head loss $g_\theta(z)=\ell(\psi_\theta(z),c)$, where $\psi_\theta$ is the output layer applied to the embedding, has an approximant $\tilde g_\theta\in H$ with $\|\tilde g_\theta\|_H\le B$ and $\sup_{\mathcal{Z}}|g_\theta-\tilde g_\theta|\le\kappa$, where $\kappa$ can be made arbitrarily small at the cost of a larger $B$ because the Gaussian kernel is universal on compact sets. Finally, for a fresh real batch, the $n_T$ real samples used in $\mathcal{L}_{\mathrm{emb}}$ are i.i.d.\ from $P$ and independent of $\Sset$ and $\theta^{\Sset}$.
 
\begin{corollary}
\label{cor:main}
Under the four assumptions above, let $\delta\in[r_0,1)$, $s=1/(1-\delta)$, $\rho_\pm=s\pm\sqrt{s^{2}-1}$, and $C(\delta)=1+\rho_+^{2}$. The displacements then satisfy $\cos\angle(G^{\Sset},G^{\T})\ge1-\delta$ and $\|G^{\Sset}\|/\|G^{\T}\|\in[\rho_-,\rho_+]$, and the parameter gap obeys the bound of Eq.~\eqref{eq:cor_a},
\begin{equation}
\|\Delta\|\le\min\Big\{\Omega\sqrt{r},\ \sqrt{\delta\,C(\delta)}\,\|G^{\T}\|\Big\},\qquad C(\delta)\xrightarrow{\delta\to0}2.
\label{eq:cor_a}
\end{equation}
The real-loss gap to the real-trained network satisfies $|L_P(\theta^{\Sset})-L_P(\theta^{\T})|\le\Lambda\|\Delta\|$ and Eq.~\eqref{eq:cor_b},
\begin{equation}
L_P(\theta^{\Sset})-L_P(\theta^{\T})\le\|\nabla L_P(\theta^{\T})\|\,\|\Delta\|+\tfrac{\beta}{2}\|\Delta\|^{2},
\label{eq:cor_b}
\end{equation}
so that if $\|\nabla L_P(\theta^{\T})\|\le\zeta$ for a constant $\zeta\ge0$, the right side is at most $\zeta\Omega\sqrt r+\tfrac{\beta}{2}\Omega^{2}r$. Moreover, for any $\eta'\in(0,1)$, with probability at least $1-\eta'$ over the real batch, Eq.~\eqref{eq:cor_c} holds,
\begin{equation}
L_P(\theta^{\Sset})\le L_{\Sset}(\theta^{\Sset})+B\Big[\sqrt{\mathcal{L}_{\mathrm{emb}}}+\tfrac{1+\sqrt{\ln(1/\eta')}}{\sqrt{n_T}}\Big]+2\kappa,
\label{eq:cor_c}
\end{equation}
where $\mathcal{L}_{\mathrm{emb}}$ is the class-wise term of Eq.~\eqref{eq:emb} at $\theta^{\Sset}$ and $L_{\Sset}(\theta^{\Sset})$ is the loss of the inner network on $\Sset_c$.
\end{corollary}
 
\begin{proof}
We first show that the normalized distance depends on the direction and scale of the two updates, then use it to bound the parameter gap, then the loss gap, and finally the MMD term.
 
Put $a=G^{\Sset}$, $b=G^{\T}$, $\rho=\|a\|/\|b\|$, and let $\varphi$ be the angle between them. Expanding $\|a-b\|^{2}=\|a\|^{2}+\|b\|^{2}-2\|a\|\|b\|\cos\varphi$ and dividing by $\|a\|^{2}+\|b\|^{2}=(1+\rho^{2})\|b\|^{2}$ gives Eq.~\eqref{eq:geom},
\begin{equation}
r_0=\frac{\|a-b\|^{2}}{\|a\|^{2}+\|b\|^{2}}=1-\frac{2\rho\cos\varphi}{1+\rho^{2}},
\label{eq:geom}
\end{equation}
where $r_0=(1+\tau)r$ by Eq.~\eqref{eq:r_tau}. Since $|\cos\varphi|\le1$ and $2\rho/(1+\rho^{2})\le1$ by the AM--GM inequality, $r_0\in[0,2]$. Cosine matching cannot see $\rho$: for $a=\lambda b$ its loss is $0$ for every $\lambda>0$, whereas Eq.~\eqref{eq:geom} equals $(1-\lambda)^{2}/(1+\lambda^{2})$, which tends to $1$ as $\lambda\to0$ or $\infty$.
Now let $r_0\le\delta<1$. Eq.~\eqref{eq:geom} gives $2\rho\cos\varphi/(1+\rho^{2})\ge1-\delta>0$, that is, Eq.~\eqref{eq:cos_lb},
\begin{equation}
\cos\varphi\ge(1-\delta)\,\frac{1+\rho^{2}}{2\rho}.
\label{eq:cos_lb}
\end{equation}
Since $(1+\rho^{2})/(2\rho)\ge1$, Eq.~\eqref{eq:cos_lb} yields $\cos\varphi\ge1-\delta$. Since also $\cos\varphi\le1$, Eq.~\eqref{eq:cos_lb} implies $\rho+1/\rho\le2s$, i.e.\ $\rho^{2}-2s\rho+1\le0$. The roots of this quadratic are $\rho_\pm=s\pm\sqrt{s^{2}-1}$ with $\rho_-\rho_+=1$, so $\rho\in[\rho_-,\rho_+]$. This proves the claims on direction and scale. 
For the parameter gap, $\theta^{\Sset}=\theta^{0}-G^{\Sset}$ and $\theta^{\T}=\theta^{0}-G^{\T}$ give $\Delta=G^{\T}-G^{\Sset}$, so $\|\Delta\|=\|a-b\|$. By the definition of $r$ in Eq.~\eqref{eq:r_tau} and the bounded-step assumption, $\|\Delta\|^{2}=r(\|a\|^{2}+\|b\|^{2}+\epsilon)\le r\,\Omega^{2}$. By Eq.~\eqref{eq:geom} and $\rho\le\rho_+$, $\|\Delta\|^{2}=r_0(1+\rho^{2})\|b\|^{2}\le\delta(1+\rho_+^{2})\|b\|^{2}$. Both bounds hold at once, which gives Eq.~\eqref{eq:cor_a}. For the limit of $C$, $\rho_+$ is a root of $\rho^{2}-2s\rho+1$, so $\rho_+^{2}=2s\rho_+-1$ and $C(\delta)=2s\rho_+$, which tends to $2$ as $\delta\to0$ because $s\to1$ and $\rho_+\to1$.
 \\
For the loss gap, convexity of $\Theta$ puts the segment from $\theta^{\T}$ to $\theta^{\Sset}$ inside $\Theta$, where $L_P$ is $\Lambda$-Lipschitz, so $|L_P(\theta^{\Sset})-L_P(\theta^{\T})|\le\Lambda\|\Delta\|$. For the one-sided bound, $\beta$-smoothness gives the descent inequality $L_P(\theta^{\Sset})\le L_P(\theta^{\T})+\langle\nabla L_P(\theta^{\T}),\Delta\rangle+\tfrac{\beta}{2}\|\Delta\|^{2}$, and the Cauchy--Schwarz inequality bounds the inner product by $\|\nabla L_P(\theta^{\T})\|\|\Delta\|$, which is Eq.~\eqref{eq:cor_b}. If $\|\nabla L_P(\theta^{\T})\|\le\zeta$, substituting $\|\Delta\|\le\Omega\sqrt r$ gives the stated form; when $\theta^{\T}$ is near-stationary the gap is therefore $O(r)$ rather than $O(\sqrt r)$.\\ 
For the MMD term, let $\theta=\theta^{\Sset}$, let $\mu_{\Sset}$ be the empirical measure of the embeddings of $\Sset_c$ at $\theta$, and let $\nu$ be the law of $\phi_\theta(x)$ with $x\sim P$. Then $L_{\Sset}(\theta)=\int g_\theta\,d\mu_{\Sset}$ and $L_P(\theta)=\int g_\theta\,d\nu$. With the kernel mean embedding $m_\mu=\int k(\cdot,z)\,d\mu(z)\in H$ of a measure $\mu$, the reproducing property and the Cauchy--Schwarz inequality give $\big|\int\tilde g_\theta\,d(\nu-\mu_{\Sset})\big|=|\langle\tilde g_\theta,m_\nu-m_{\mu_{\Sset}}\rangle_H|\le B\,\|m_\nu-m_{\mu_{\Sset}}\|_H$. Because $|g_\theta-\tilde g_\theta|\le\kappa$ on $\mathcal{Z}$, replacing $\tilde g_\theta$ by $g_\theta$ changes each integral by at most $\kappa$, hence Eq.~\eqref{eq:rkhs},
\begin{equation}
L_P(\theta)-L_{\Sset}(\theta)\le B\,\|m_\nu-m_{\mu_{\Sset}}\|_H+2\kappa.
\label{eq:rkhs}
\end{equation}
Let $\hat\nu$ be the empirical measure of the $n_T$ real embeddings. Expanding the squared RKHS norm into three kernel averages shows that $\|m_{\mu_{\Sset}}-m_{\hat\nu}\|_H^{2}$ equals the class-wise term of Eq.~\eqref{eq:emb}, so the triangle inequality gives $\|m_\nu-m_{\mu_{\Sset}}\|_H\le\sqrt{\mathcal{L}_{\mathrm{emb}}}+F$ with $F=\|m_{\hat\nu}-m_\nu\|_H$. It remains to bound $F$. Given $\theta^{\Sset}$, the real embeddings $z_1,\dots,z_{n_T}$ are i.i.d.\ with law $\nu$, so $\mathbb{E}F^{2}=\tfrac{1}{n_T}\big(\mathbb{E}\,k(z,z)-\|m_\nu\|_H^{2}\big)\le1/n_T$, because $k(z,z)=1$, and Jensen's inequality gives $\mathbb{E}F\le1/\sqrt{n_T}$. Replacing a single $z_i$ by $z_i'$ changes $F$ by at most $\tfrac{1}{n_T}\|k(\cdot,z_i)-k(\cdot,z_i')\|_H=\tfrac{1}{n_T}\sqrt{2-2k(z_i,z_i')}\le\sqrt2/n_T$, since $k>0$. McDiarmid's inequality then gives $\Pr[F\ge\mathbb{E}F+t]\le\exp\!\big(-2t^{2}/(n_T\cdot2/n_T^{2})\big)=\exp(-n_Tt^{2})$. Setting the right side to $\eta'$, i.e.\ $t=\sqrt{\ln(1/\eta')/n_T}$, yields $F\le(1+\sqrt{\ln(1/\eta')})/\sqrt{n_T}$ with probability at least $1-\eta'$. Inserting this bound into Eq.~\eqref{eq:rkhs} gives Eq.~\eqref{eq:cor_c} and completes the proof.
\end{proof}
 
\noindent The corollary shows that this bounds the real-loss gap to the real-trained network, and that the MMD term bounds the real loss of the synthetic-trained network by its synthetic loss, which is why $\lambda_{\mathrm{emb}}=1$ balances two quantities bounded on comparable scales. The constants $\Lambda,\beta,B,\kappa$ are not measured, so the bounds are qualitative, and the corollary concerns the real training distribution $P$ rather than unseen subjects.

 